%% file: root.tex
\documentclass[letterpaper, 10 pt, journal, twoside]{IEEEtran}
\pdfoutput=1
\input{tikz_styles}

\usepackage{graphicx} %
\usepackage{subcaption} %
\usepackage{amsmath} %
\usepackage{amssymb}  %
\usepackage{multirow}
\usepackage{cite}
\usepackage{algorithmic}
\usepackage{xcolor}
\usepackage{moreverb,url}
\usepackage{dsfont}
\usepackage{comment}
\usepackage{subfloat}

\usepackage{hyperref} %
\usepackage[ruled,vlined,noresetcount]{algorithm2e}

\usepackage{pdfcomment}
\usepackage{environ}
\RenewEnviron{comment}{\pdfcomment{\BODY}}
\begin{document}
\title{Why-So-Deep: Towards Boosting Previously Trained Models for Visual Place Recognition}

\author{M.~Usman~Maqbool~Bhutta$^{1}$, Yuxiang~Sun$^{2}$, Darwin~Lau$^{3}$, and Ming~Liu$^{4}$%
	\thanks{Manuscript received: September 9, 2021; Revised October 21, 2021; Accepted December 20, 2021.}
	\thanks{This paper was recommended for publication by 		Editor Sven Behnke upon evaluation of the Associate Editor and Reviewers’ comments. \textit{(Corresponding Author: M. Usman Maqbool Bhutta)}
	} 
	\thanks{$^{1}$M. Usman Maqbool Bhutta and $^{2}$Darwin Lau are with the C3 Robotics Lab, Department of Mechanical and Automation, CUHK, Hong Kong.
		{\tt\footnotesize usmanmaqbool@outlook.com ; darwinlau@cuhk.edu.hk}}%
	\thanks{$^{2} $Yuxiang Sun is with the Department of Mechanical Engineering, PolyU, Kowloon, Hong Kong.
		{\tt\footnotesize sun.yuxiang@outlook.com}}
	\thanks{$^{4} $Ming Liu is with the Department of Electronic and Computer Engineering, HKUST, Hong Kong.
		{\tt\footnotesize eelium@ust.hk}}
	\thanks{Digital Object Identifier (DOI): see top of this page.}
}

\markboth{IEEE Robotics and Automation Letters. Preprint Version. Accepted January, 2022}
{Bhutta \MakeLowercase{\textit{et al.}}: Why-So-Deep}

\IEEEtitleabstractindextext{%
\begin{abstract}
Deep learning-based image retrieval techniques for the loop closure detection demonstrate satisfactory performance. However, it is still challenging to achieve high-level performance based on previously trained models in different geographical regions. This paper addresses the problem of their deployment with simultaneous localization and mapping (SLAM) systems in the new environment. The general baseline approach uses additional information, such as GPS, sequential keyframes tracking, and re-training the whole environment to enhance the recall rate. We propose a novel approach for improving image retrieval based on previously trained models. We present an intelligent method, \textit{MAQBOOL}, to amplify the power of pre-trained models for better image recall and its application to real-time multiagent SLAM systems. We achieve comparable image retrieval results at a low descriptor dimension (512-D), compared to the high descriptor dimension (4096-D) of state-of-the-art methods. We use spatial information to improve the recall rate in image retrieval on pre-trained models. Material related to this work is available at \url{https://usmanmaqbool.github.io/why-so-deep}.

\vspace{0.15cm}
 
\end{abstract}


}
\maketitle
\IEEEdisplaynontitleabstractindextext

\IEEEpeerreviewmaketitle
\section{Introduction}
\label{sec.relatedwork}
\IEEEPARstart{V}{isual} place recognition is extensively used in the robotics industry with application in all kinds of SLAM systems for the loop closure detection \cite{qin2018vins, Leutenegger2013}. The top candidate in the image retrieval helps a lot in the multiagent SLAM system for creating large-scale 3D maps \cite{Schneider2018,bhutta2020loopbox,Bhutta2018}. Convolutional neural network (CNN)-based approaches such as NetVLAD \cite{arandjelovic2016netvlad} and DGC-Net \cite{melekhov2019dgc} produce promising results in the image retrieval. 

There are several types of common solutions to enhance the localization performance further. Each of the methods involves retraining the whole network in addition to the data. %
The first and very common approach in visual place recognition is based on attention-seeking regions in the images. Work-related to landmarks distribution and partitioning of pictures based on regions is also present in the literature \cite{Panphattarasap2017, xin2019localizing, Sunderhauf2015, spatialpaperreview}. \cite{Zhai} incorporated NetVLAD and made a features pyramid of the top landmarks in the image to achieve better results. \cite{Chen2017} introduced a multi-layered region-based method that extends DGC-Net and incorporates NetVLAD. Even though the better real-time performance of DGC-Net \cite{melekhov2019dgc}, sometimes, the system faces difficulty due to the incorporation of objects changing with time in the images. 
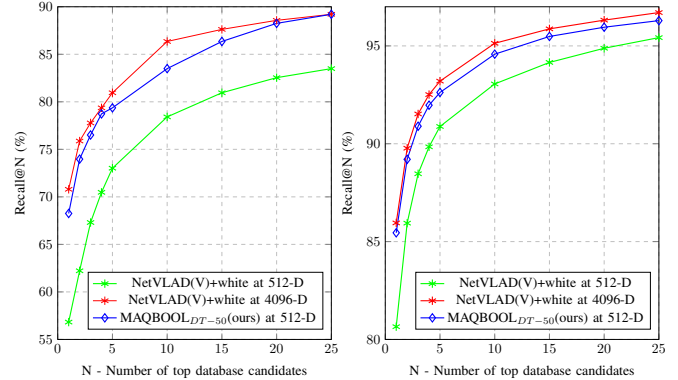
\begin{figure}[!t] 
	\centering
	\raggedright	
	\captionsetup[subfigure]{width=.85\linewidth}
	\subfloat[Recall test on Tokyo 24/7] 
	{	\label{fig:pitts2tokyo-512} 
		\resizebox{.45\linewidth}{!}{
			\pgfplotsset{ymax=90, ymin= 55}
			\begin{tikzpicture}[trim axis left]%
				\begin{axis}
					[curve plot style]
					\addplot [color=green, mark=asterisk, mark size=3pt, line width=0.8pt]
					table {data/vd16_tokyoTM_to_tokyo247_netvlad_512.dat};
					
					\addplot [color=red, mark=asterisk, mark size=3pt]
					table {data/vd16_tokyoTM_to_tokyo247_netvlad_4096.dat};
					\addplot [color=blue, mark=diamond, mark size=3pt, line width=0.8pt]
					table {data/vd16_tokyoTM_to_tokyo247_maqbool_DT_50_512.dat};
					\legend{NetVLAD(V)+white at 512-D, NetVLAD(V)+white at 4096-D, MAQBOOL$_{DT-50}$(ours) at 512-D}
					
				\end{axis}
			\end{tikzpicture}
		}
	}
	\captionsetup[subfigure]{width=.85\linewidth}
	\subfloat[Recall test on Pitts250k] 
	{\label{fig:pitts2tokyo-4096}
		\resizebox{.45\linewidth}{!}{
			\pgfplotsset{ymax=97, ymin= 80}
			\begin{tikzpicture}[trim axis left]%
				\begin{axis}
					[curve plot style]
					\addplot [color=green, mark=asterisk, mark size=3pt, line width=0.8pt]
					table {data/vd16_pitts30k_to_pitts30k_netvlad_512.dat};
					\addplot [color=red, mark=asterisk, mark size=3pt, line width=0.8pt]
					table {data/vd16_pitts30k_to_pitts30k_netvlad_4096.dat};
					\addplot [color=blue, mark=diamond, mark size=3pt, line width=0.8pt]
					table {data/vd16_pitts30k_to_pitts30k_maqbool_DT_50_512.dat};
					
					\legend{NetVLAD(V)+white at 512-D,NetVLAD(V)+white at 4096-D, MAQBOOL$_{DT-50}$(ours) at 512-D }
				\end{axis}
			\end{tikzpicture}
		}
	}
	\caption{Our system performance at a lower feature dimension (512-D) is comparable to NetVLAD  \cite{arandjelovic2016netvlad} performance at a higher feature dimension (4096-D). We use NetVLAD best trained network $(VGG-16,f_{VLAD})$ for evaluation.}
	\label{fig:a}
\end{figure}

The second most used idea is to increase the depth of the neural network by adding several additional layers with the default network. For instance, a multi-layered region-based framework is introduced in \cite{laskar2020geometric} which also uses NetVLAD and performs dense pixel matching to achieve better place recognition. For large-scale image correspondence matching, scientists have also introduced several upgrades to the network. \cite{Yan2015} presented HD-CNN, a hierarchical deep CNN scheme, while \cite{Yu2019SpatialRecognition} and \cite{camara2019spatio} similarly partitioned the spatial information in higher layers. But the utilization of these methods in a real-time SLAM system is still very challenging due to the computation time and large size of feature dimension. 

The third widely used method in the robotics industry is done by integrating 3D-depth information. The corresponding 2-D images, along with the 3D maps, enhance the system performance in loop closure detection \cite{Sarlin2018, Gawel2018, bernreiter2019multiple,taira2018inloc}. Moreover, GPS information \cite{pillai2019self}, semantics graph matching \cite{Gawel2018}, and attention-seeking approaches \cite{xin2019localizing, Chen2017, Pal2019} have shown good results for image retrieval tasks. In robotics, the place recognition module should not be tightly coupled with GPS or 3D data. Otherwise, it will become harder for the server of the multiagent SLAM system to handle. Despite their benefits, all the above-described approaches require retraining the complex networks to enhance system accuracy. 

In our proposed study, our system at a lower feature dimension (i.e., 512-D) is able to achieve accuracy similar to a higher feature dimension (i.e., 4096-D) while tested with the same pre-trained model. A detailed comparison is shown in Fig \ref{fig:a}. In our work, instead of following a cascading training pipeline or to go deeper in terms of feature dimension, we probabilistically enhance the image retrieval performance based on a pre-trained model for a more reliable place recognition. So, we introduce this as the Multiple AcQuisitions of perceptiBle regiOns for priOr Learning (MAQBOOL) approach. These significant correspondence regions will help in probabilistic landmarks elevation by splitting the full spatial information into multiple regions and estimate their descriptor $\ell_2$ distance co-relations, which significantly increases the power of the pre-trained models. In addition, training new models to be utilized in new places involves intensive computation of deep learning models each time.

As multiagent SLAM is the next interest for computer vision researchers, we strongly believe that our contributions will help the computer vision community in many multiagent SLAM scenarios. The contribution of this research is four-fold, as follows:
\begin{itemize}
	\item Our main contribution is to enhance recall accuracy of previously trained models.
	\item We introduce a probabilistic layer to constrain the local representation such as prominent regions, along with the global representation of images. The global description of the pre-trained model and the local consistency introduced in our work enables the system to yield good performance at a large-scale.	
	\item Our system shows comparable accuracy at a low descriptor dimension (512-D) compared to the high descriptor dimension (4096-D) of the current state-of-the-art \cite{arandjelovic2016netvlad}. 
	\item In our results, we show good performance of our system at low-dimensional features with a previously trained model while tested in the new environment. It enables the SLAM system to detect the loop closure at good accuracy everywhere.
\end{itemize}

\begin{figure*}[!t]
	\centering
	
	\includegraphics[width=0.95\textwidth]{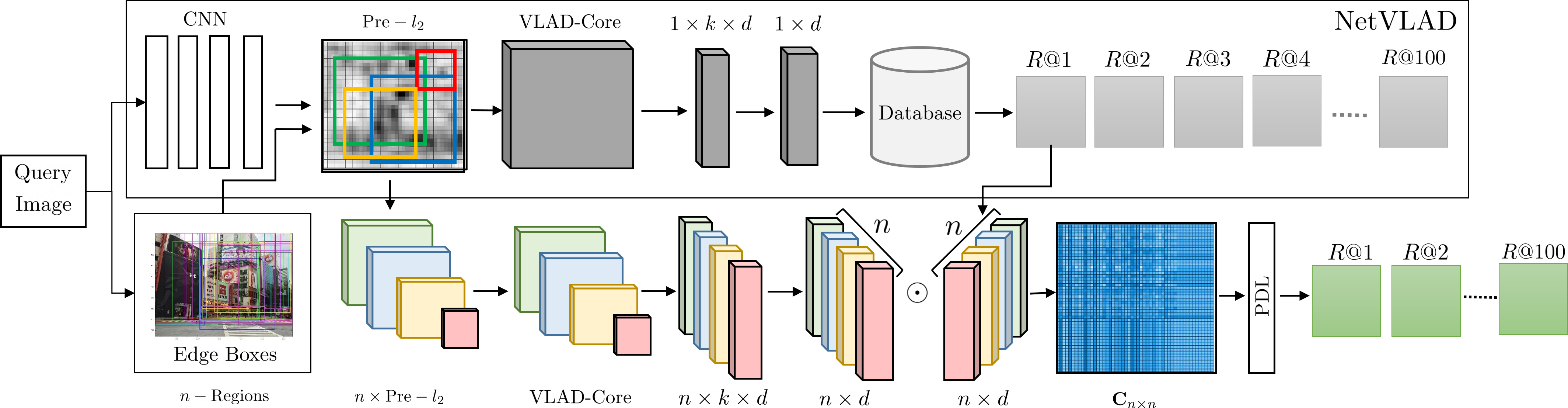}
	\caption{Overview of the MAQBOOL system. For each image, the spatial $\text{Pre}-l_2$ layer is partitioned according to the boxes detected by edge boxes. All the regions are cropped and then normalized for the feature vectors. $k$ is the number of clusters, and $d$ is the feature dimension. The probabilistic decision layer (PDL) is trained using the correlation distance matrix $\mathbf{C}_{n \times n}$  and the ground truth of the ToykoTM dataset.}
	\label{figure-main}
\end{figure*} 

\section{Related works}
\label{sec.relatedwork}
This work is related to improving image retrieval, which plays a big part in visual place recognition for vision-related applications and any kind of SLAM system. To improve image retrieval, scientists have worked towards making robust global descriptors, and some have utilized local features along with a global representation, as mentioned in previous section.
We discuss closely related work below.

Researchers developed a geometric image correspondence-based system, which shows good performance after utilizing dense geometric information \cite{laskar2020geometric}. It selects the top candidates from the database using NetVLAD and performs dense pixel matching with the query image. Their geometric model was created by fitting the planer homography to the 3D information, SIFT features and CNN descriptor. This was done with alteration to DGC-Net \cite{melekhov2019dgc}, which does not deal with the 3D structure. After modifying DGC-Net, \cite{laskar2020geometric} utilized a unified correspondence map decoder (UCMD) for the dense matching between the top candidates and the query image. It processes a multi-resolution feature pyramid and CNN layers. At the end, the authors used a neighbourhood consensus networks (NCNet), meaning we can summarize their system as NetVLAD-DGC-NC-UCMD-Net. This long computation pipeline makes their system complex, and it requires intense computation for each query image and all database images.

\cite{taira2018inloc} also considered 3D information and performed extensive nearest-neighbour explorations in the descriptive space. In addition, they used coarse correspondence estimation, while \cite{laskar2020geometric} used a learned convolutional decoder. Another pyramid-aggregation-based method is found in \cite{Yu2019SpatialRecognition}. It also uses previous NetVLAD training results along with enhancing the accuracy compared to NetVLAD, and introduces weighted triplet loss for updating the weights after each epoch while training the new model.

\cite{zhu2018attention} introduced APANet, which uses principal component analysis (PCA) power whitening along with pyramid aggregation of attentive regions. APANet selects the features of key attention-seeking regions and performs sum pooling, and its results are on 512-D features trained using the AlexNet \cite{alexnet} and VGG-16 \cite{VGG-network} networks. Their work is similar to \cite{laskar2020geometric} for multi-scale region aggregation to build the pyramid. In addition, their use of PCA power whitening makes it more complex for the system to create a full descriptor and inconvenient to use in real-time applications.

The training-free approach from \cite{Sunderhauf2015} uses edge boxes \cite{Zitnick2014} for the top landmarks selection in a given scene. This method scales down each landmark feature from a 65K vector to a 1K-dimensional representation using Gaussian random projection. Based on the landmark features, cosine distances between all the landmark proposals are calculated to find the similarity. No training is involved, for better utilization of the landmarks. An unsupervised approach \cite{reviewerpaper} presents a method to re-ranking the NetVLAD top-20 candidates. Their method utilized global as well as local features for improving the recall rate.

\section{Multiple acquisitions of perceptible regions for prior learning}
\label{sec.framworkwork}

An overview of our MAQBOOL system is shown in Fig. \ref{figure-main}. Given a query image, we retrieve the nearest candidates in the database descriptor space. We use NetVLAD, which is a fast and scalable method, for the image retrieval application. 

Our proposed method consists of three parts. Part I explains the selection of the top regions for the local representation from the images; Part II describes the corresponding spatial information processing using NetVLAD; and Part III shows the probabilistic manipulation, in a pairwise manner, of the query image with initially retrieved database images for more reliable place recognition. These three parts are explained in the following subsections.

\subsection{Top Regions Selection}
In Section \ref{sec.relatedwork}, we explained that APANeT \cite{zhu2018attention} uses the top regions and applies single and cascaded blocks to achieve better performance. After taking inspiration from the top regions selection in \cite{Sunderhauf2015}, we also choose edge boxes \cite{Zitnick2014} instead of using the simple regions selection techniques from \cite{xiao2015application-47} and \cite{yan2017fine-48}. Edge boxes is a fast and unsupervised method that detects the likelihood of an object based on the points on the edges. 

\subsection{Spatial Landmarks Estimation }
\begin{figure}[!t]
	
	\begin{subfigure}[t]{.5\textwidth}
		\centering
		\includegraphics[width=.97\textwidth]{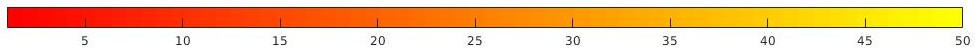}
	\end{subfigure}%
	
	\vspace{.1cm}
	\begin{subfigure}[t]{0.25\textwidth}
		\centering
		\captionsetup{width=.9\linewidth}
		\includegraphics[width=.9\textwidth]{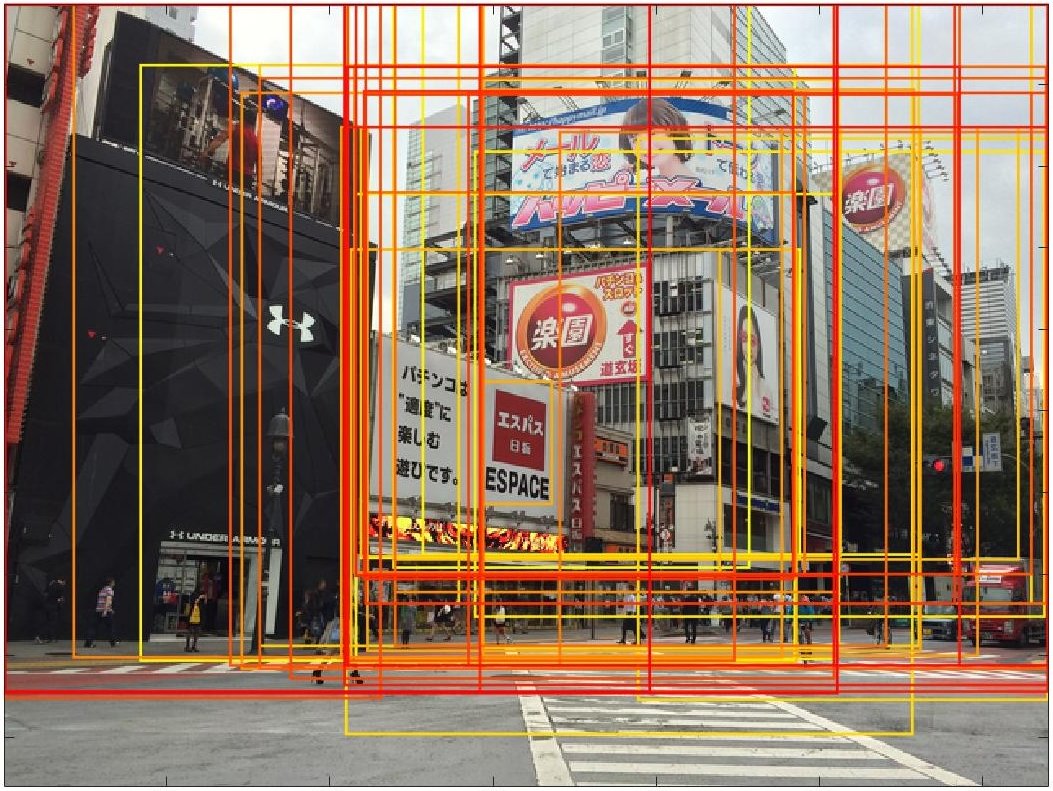}
		\caption{Top-ranked image correspondence proposals are marked on the query image.}
		\label{fig:first-1}
	\end{subfigure}%
	\begin{subfigure}[t]{0.25\textwidth}
		\centering
		\captionsetup{width=.9\linewidth}
		\includegraphics[width=.91\textwidth]{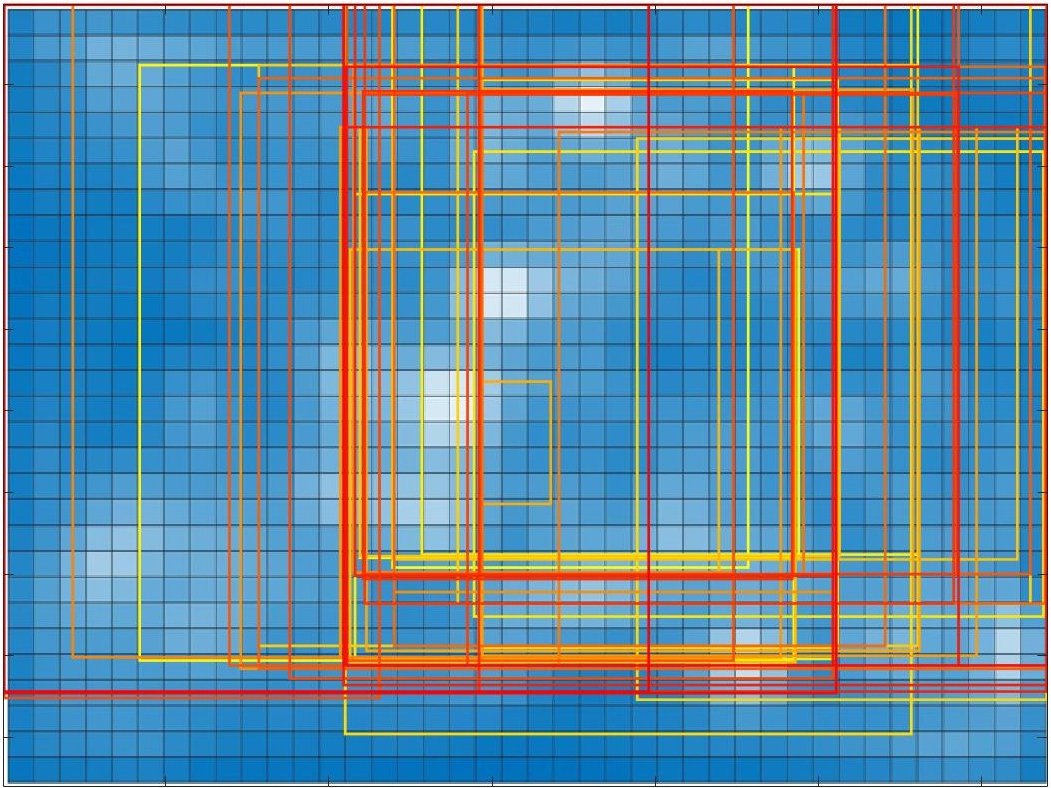}
		\caption{Proposals are mapped to the spatial information of pre-$\ell_2$ layer of NetVLAD}
		\label{fig:first-2}
	\end{subfigure}%
	\caption{Query image from the Tokyo 24/7 dataset. Top region proposals are selected using edge boxes \cite{Zitnick2014}. The intensity bar shows the top-scoring region selections on the image.}
	\label{fig:first}
	
\end{figure}
We process the landmarks information at the pre-$\ell_2$ layer of NetVLAD. All the top regions' enclosed boxes are mapped to this pre-$\ell_2$ layer, and we crop the corresponding spatial information to create the NetVLAD 512-D and 4096-D feature vectors of all the boxes. These feature vectors are used for the probabilistic landmarks elevation in the next section.
Essential regions are calculated in the query image and candidate images from the database. Fig. \ref{fig:first} shows the top proposals based on the objectness scores and their mapping at the pre-$\ell_2$ layer of the NetVLAD pipeline.  

For a query image $\textbf{I}_q$, NetVLAD predicts top-100 matches from the database, based on their $\ell_2$ distances in the descriptor space. Let's assume, for this query image $\textbf{I}_q$, $\mathbf{I}_{c}$ are the top candidates from the database and $d_c^j$ is the corresponding $\ell_2$ feature distance of $j^{th}$ candidate image from the query image. We further denote the feature distance $^qd_{crn}^j$ between query and different regions of $j^{th}$ candidate image pictures by a double-sided solid arrow, as shown in Fig. \ref{fig-method}. We calculate relative distances $R^\prime$ of two adjacent candidate proposals by taking the derivative of the original distance set $d_c$.

Instead incorporating multi-scale pyramid aggregation or estimating sets of cyclically consistent dense pixel matches, we simply performed probabilistic spatial landmarks elevation followed by the correlation distance for the re-ranking of the top retrieved candidates. For the $d$-dimensional feature vector and $n$ (number of spatial regions plus one considering the full image), we define the whole image representation by \textbf{F}, which has a size of $n \times d$. We estimate the correlation distance matrix $\textbf{C}_{n \times n}$ of the query image $\textbf{I}_q$ with the $j^{th} \in \mathbf{I}_{c}$ database images $\textbf{I}_{c}^j$ as

\begin{equation} 
	\textbf{C}_{n \times n}^j = \textbf{F}_{q(n \times d)} 
	\times {\textbf{F}^j_{c(n \times d)}}^T,
\end{equation}
We filter the irrelevant distances score from $\textbf{C}_{n \times n}$ after subtracting the maximum distances $d_{c}^{max} \in d_{c}$ and using sign function:
\begin{figure}[!t]
	\centering
	\includegraphics[width=0.48\textwidth]{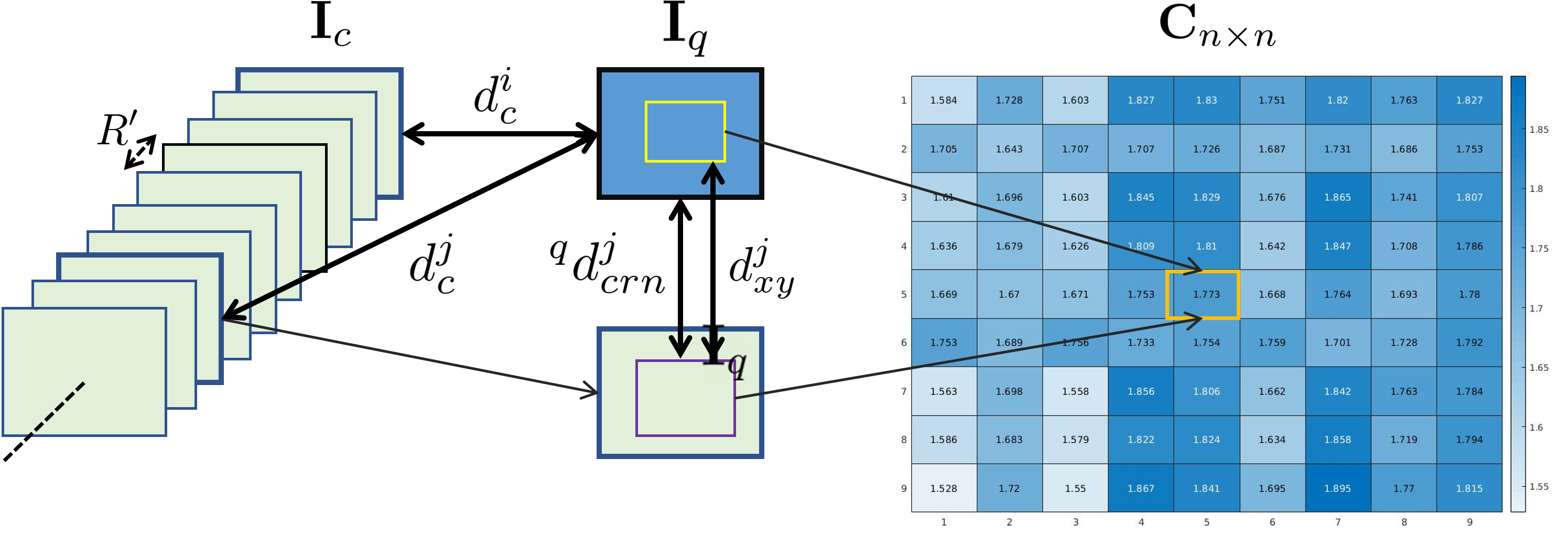}
	\caption{The query image is compared with all the top candidates. $R^\prime$ is the distance span within adjacent image proposals. The correlation distance matrix $\textbf{C}_{n \times n}$ is calculated between the query image and the candidate images by taking the feature distance between all top landmarks. Each value of this matrix represents the feature distance between specific landmarks shown by a yellow box, and their distance is represented as $d_{xy}^j$.}
	\label{fig-method}
\end{figure}
\begin{equation}   \textbf{C}_{f}^j =  \textbf{C}_{n \times n}^j - d_{c}^{max}.\end{equation} 
\begin{equation} 
	\textbf{D}^{j} = 
	\begin{cases}
		sgn(\textbf{C}^j_f), & \text{for } c_{f} < 0 \\
		0, & \text{for } c_{f} \geq 0, \\
	\end{cases}
\end{equation} 
where $c_{f} \in \textbf{C}_{f}^j$. By using $ \textbf{D}^{j} $, we drop large distances landmarks from $\textbf{C}_{n \times n}^j$ as follows:

\begin{equation}  \textbf{C}_{n \times n}^j = |\textbf{D}_{qj}^j| \odot \textbf{C}_{n \times n}^j.
\end{equation} 

\subsection{Probabilistic Spatial Landmarks Elevation}

Firstly, we determine the information $\textbf{S}^j$ based on the size of the boxes of both the query and database image:

\begin{equation} 
	s_{xy}^j = \beta* P_b^q* P_b^{c} * e^{-(d_{c}^{min}+d_{xy}^j)} * e^{-R_j^\prime},
\end{equation} 
\\
where $s_{xy}^j \in \textbf{S}^j$, $x,y \in \{1,...n\}$, $\beta=10$, $d_{xy}^j \in \textbf{C}_{n \times n}^j$, and $R_j^\prime = \frac{d}{dx}(d_{c}^j)$ is the candidates' relative distances' difference. Furthermore, $b_w$ and $b_h$ are the width and height, respectively, of the landmarks' bounding box. We estimate the probabilities of the landsmarks in the query and candidate images as follows:
\begin{equation} 
	P_b^q = e^{-\frac{b_w^q*b_h^q}{q_w*q_h} } \text{ and } P_b^{c} = e^{-\frac{b_w^{c}*b_h^{c}}{c_w*c_h} }.
\end{equation} 
The reason for taking the negative exponential of the weight of the landmarks' over the full image is that a smaller bounding box results in a lower probability of a good match. For instance, in night images, most regions are black, so there exists a higher chance of accumulating dark region effects.

We use information matrix $\textbf{S}^j$ and filtered correlation distance matrix $\textbf{C}_f^j$ to produce the probabilistic correlation matrix $\textbf{P}^j_{SC}$ as follows:

\begin{equation} 
	\textbf{P}^j_{SC} =  \textbf{S}^j \odot \textbf{C}_f^j. 
\end{equation} 
Each column of $\textbf{P}^j_{SC}$ corresponds to a probabilistic match of a particular landmark in the query image with all landmarks in the $jth$ candidate image. We further shrink $\textbf{P}^j_{SC}$ to $10 \times 10$ by sorting each column of $\textbf{C}_{n \times n}$ and index matching with $\textbf{P}^j_{SC}$.

$ P^j_{SM}$ processes the original $\ell_2$ distances of all retrieved candidates from the query image. We use softmax for estimating the probability controlled by the $c_{min}$:

\begin{equation} 
	P^j_{SM} = e^{-c_{min}^j}.\underbrace{\left(\frac{e^{-d_{c}^j}}{ \sum_i e^{-d_{c}^i}}\right)}_\text{softmax}=e^{-c_{min}^j}.\text{softmax}(e^{-d_{c}^j}),
\end{equation} 
where $c_{min}^j$ is the minimum value of $\textbf{C}_{n \times n}^j$. We pass this estimated probability information to the regression process to create the predictive model.

We estimate the probability $\textbf{M}^j$  by utilizing the $P^j_{SM}$ and $\textbf{P}^j_{SC}$: 
\begin{equation} 
	\textbf{M}^j =  P^j_{SM} * \textbf{P}^j_{SC}.
\end{equation} 

We also consider incorporation of the feature distance $d_{c}^j$ between the query image and the candidate image, as well as the distances $^qd_{crn}^j$ between the top regions of the candidate image, with the whole query image, as follows

\begin{equation} 
	\textbf{C}_{qc}^j = [R_j^\prime, ^qd_{cr1}^j, ^qd_{cr2}^j, ^qd_{cr3}^j,...,^qd_{crn}^j  ],
\end{equation}

In this work, we choose the top 10 feature distances $^qd_{crn}^j$ of the candidate image's top regions from the whole query image.

\subsection{Prediction Model and Probabilistic Distance Update}

We design the probabilistic decision layer (PDL) using the ground truth based on the TokyoTM validation dataset. The information estimated in the previous subsection is used in creating the model as follows:

\begin{equation}  P_M^j =  f(d_{c}^j, \textbf{C}_{qc}^j , \textbf{M}^j, Y^j). \end{equation} 

We train $P_M^j$ with the ground truth $Y^j$ of about 250 images and create the model. We choose bootstrap aggregation in the decision tree (DT), which allows the tree to grow on an independently drawn bootstrap, duplicate of the input. This reduces the variance and increases accuracy. We trained bootstrap-aggregated decision trees of sizes 50 (DT-50) and 100 (DT-100) for the testing. After creating the model, we apply it to work like the prior distribution to predict the response. In this manner, we update all the distances of the top candidates for re-ranking the retrieval images as follows:
\begin{equation} 
	d_{new}^j = |d_{c}^j-\alpha \log^{(P_M^j)}|, P_M^j \in [1,2].
\end{equation} 

We keep the model response binary. The predicted value '1' corresponds to an irrelevant match, while '2' indicates the nearest match. The regularizing variable $\alpha$ controls the weight of the probabilistic response. We choose $\alpha=1.15$ while working at the 512-D feature vectors, while at the higher dimension, i.e., 4096-D, we use $\alpha=0.31$. The main motivation for using $\alpha=0.31$ is to minimize the effect of the regularizing variable. We observed that the feature distance between two adjacent images is small at a higher dimensional feature space. So the impact of the regularizing variable should also be low for the high-dimensional features.

\section{Results and Discussion}
\label{sec.results}
Our proposed method requires small training of decision tree model, and it works excellently when tested in new surroundings. All the testing results in this section are based on the same decision tree model. We used the TokyoTM validation dataset to train this decision model for the PDL layer. MAQBOOL probabilistically elevates the perceptible regions' distributions for improving the loop closure module of the multiagent SLAM system. Our system performs better image retrieval than schemes that change the network with a complex structure and include additional sensor information or perform repetitive training to get impressive results on challenging datasets.
\begin{figure*}[h!t] 
	\centering
	\raggedright	
	\captionsetup[subfigure]{width=.9\linewidth}
	\subfloat[Pitts250k at 512-D] 
	{	\label{fig:4plots-pitts2pitts-512} 
		\resizebox{.225\linewidth}{!}{
			\pgfplotsset{ymax=97, ymin= 80}
			\begin{tikzpicture}[trim axis left]%
				\begin{axis}
					[curve plot style]
					\addplot [color=blue, mark=diamond, mark size=3pt, line width=0.8pt]
					table {data/vd16_pitts30k_to_pitts30k_maqbool_DT_50_512.dat}; %
					\addplot [color=green, mark=square, mark size=3pt, line width=0.8pt]
					table {data/vd16_pitts30k_to_pitts30k_maqbool_DT_100_512.dat}; %
					\addplot [color=black, mark=asterisk, mark size=3pt, line width=0.8pt]
					table {data/vd16_pitts30k_to_pitts30k_netvlad_512.dat};
					\addplot [color=magenta, mark=triangle, mark size=3pt, line width=0.8pt]
					table {data/vd16_pitts30k_to_pitts30k_apanet_512.dat};
					\legend{MAQBOOL$_{D-50}$(ours),MAQBOOL$_{D-100}$(ours), NetVLAD(V)+white,  APANet(V)}
					
				\end{axis}
			\end{tikzpicture}
		}
	}
	\captionsetup[subfigure]{width=.9\linewidth}
	\subfloat[Pitts250k at 4096-D] 
	{	\label{fig:4plots-pitts2pitts-4096} 
		\resizebox{.225\linewidth}{!}{
			\pgfplotsset{ymax=97, ymin= 85.5}
			\begin{tikzpicture}[trim axis left]%
				\begin{axis}
					[curve plot style]
					\addplot [color=blue, mark=diamond, mark size=3pt, line width=0.8pt]
					table {data/vd16_pitts30k_to_pitts30k_maqbool_DT_50_4096.dat}; %
					\addplot [color=green, mark=square, mark size=3pt, line width=0.8pt]
					table {data/vd16_pitts30k_to_pitts30k_maqbool_DT_100_4096.dat}; %
					
					\addplot [color=black, mark=asterisk, mark size=3pt, line width=0.8pt]
					table {data/vd16_pitts30k_to_pitts30k_netvlad_4096.dat};
					\legend{MAQBOOL$_{D-50}$(ours), MAQBOOL$_{D-100}$(ours), NetVLAD(V)+white}
				\end{axis}
			\end{tikzpicture}
		}
	}
	\captionsetup[subfigure]{width=.9\linewidth}
	\subfloat[Tokyo 24/7 at 512-D] 
	{\label{fig:4plots-tokyo2tokyo-512}
		\resizebox{.225\linewidth}{!}{
			\pgfplotsset{ymax=90, ymin= 56}
			\begin{tikzpicture}[trim axis left]%
				\begin{axis}
					[curve plot style]
					\addplot [color=blue, mark=diamond, mark size=3pt, line width=0.8pt]
					table {data/vd16_tokyoTM_to_tokyo247_maqbool_DT_50_512.dat}; %
					\addplot [color=green, mark=square, mark size=3pt, line width=0.8pt]
					table {data/vd16_tokyoTM_to_tokyo247_maqbool_DT_100_512.dat}; %
					
					\addplot [color=black, mark=asterisk, mark size=3pt, line width=0.8pt]
					table {data/vd16_tokyoTM_to_tokyo247_netvlad_512.dat};
					\addplot [color=magenta, mark=triangle, mark size=3pt, line width=0.8pt]
					table {data/vd16_tokyoTM_to_tokyo247__APA_512.dat};
					\legend{MAQBOOL$_{D-50}$(ours), MAQBOOL$_{D-100}$(ours),  NetVLAD(V)+white, APANet(V)}
				\end{axis}
			\end{tikzpicture}
		}
	}
	\captionsetup[subfigure]{width=.9\linewidth}
	\subfloat[Tokyo 24/7 at 4096-D] 
	{\label{fig:4plots-tokyo2tokyo-4096}
		\resizebox{.225\linewidth}{!}{
			\pgfplotsset{ymax=91, ymin= 70}
			\begin{tikzpicture}[trim axis left]%
				\begin{axis}
					[curve plot style]
					\addplot [color=blue, mark=diamond, mark size=3pt, line width=0.8pt]
					table {data/vd16_tokyoTM_to_tokyo247_maqbool_DT_50_4096.dat}; 
					\addplot [color=green, mark=square, mark size=3pt, line width=0.8pt]
					table {data/vd16_tokyoTM_to_tokyo247_maqbool_DT_100_4096.dat}; 
					\addplot [color=black, mark=asterisk, mark size=3pt]
					table {data/vd16_tokyoTM_to_tokyo247_netvlad_4096.dat};
					
					\legend{MAQBOOL$_{D-50}$ (ours), MAQBOOL$_{D-100}$ (ours), NetVLAD(V)+white}
				\end{axis}
			\end{tikzpicture}
		}
	}
	\caption{Recall @ 1-25 on Toyko 24/7 and Pittsburgh 250k datasets. (a) and (b) show the performance of MAQBOOL compared to NetVLAD based on the model pre-trained on VGG Pittsburgh 250k. (c) and (d) show the performance based on the model pre-trained on VGG-16.}
	\label{fig:4plots}
\end{figure*}
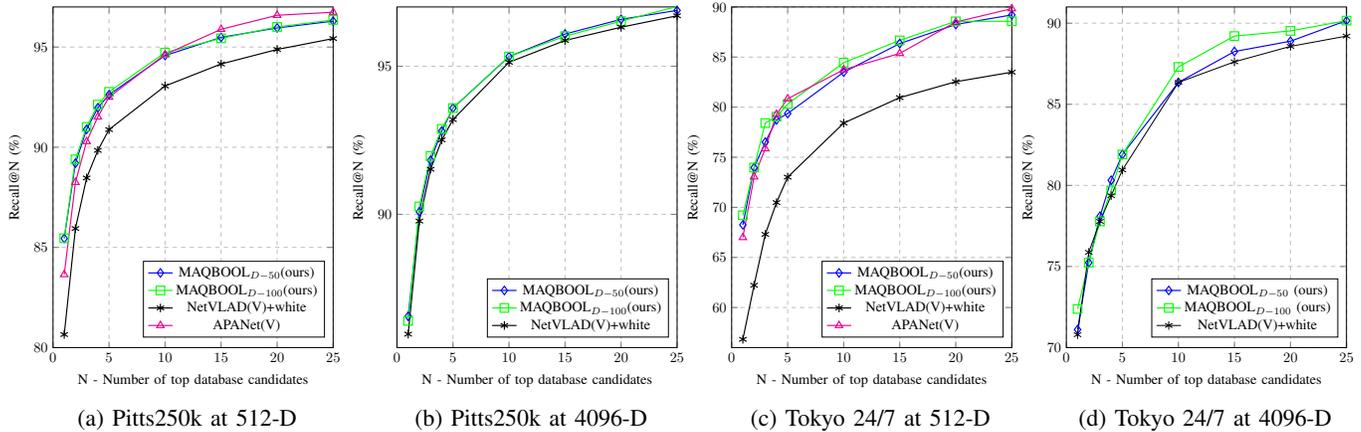

\subsection{Datasets and Implementation}
NetVLAD is mainly evaluated on  Pittsburgh \cite{Torii2015} and Tokyo 24/7 \cite{Torii2018} datasets. We used the same NetVLAD VGG-16-based models for the performance evaluation. We tested our MAQBOOL method on the Pittsburgh and Tokyo 24/7 datasets compared with NetVLAD and APANet \cite{zhu2018attention}. The Pittsburgh 250K dataset consists of 254K perspective images taken from 10.6K Google Street View and 8.2K query images, while the Tokyo 24/7 dataset has 76K database images and 315 query images. 

\subsection{Testing on Pittsburgh and Tokyo 24/7 Datasets}
Fig. \ref{fig:4plots} shows the comparison of our proposed MAQBOOL strategy while testing on the Pittsburgh and Tokyo 24/7 datasets. In comparison to the feature dimension of 4096, we achieve significant improvement at a feature dimension of 512, as shown in Fig. \ref{fig:4plots-pitts2pitts-512} and \ref{fig:4plots-tokyo2tokyo-512}. 

\begin{figure*}[!t] 
	\raggedright	
	\captionsetup[subfigure]{width=.93\linewidth}
	\subfloat[Recall test on Tokyo 24/7 dataset at a features dimension of 512.] {\label{fig:results-thumb-tokyo2tokyo-512} 
		\includegraphics[width=.99\linewidth]{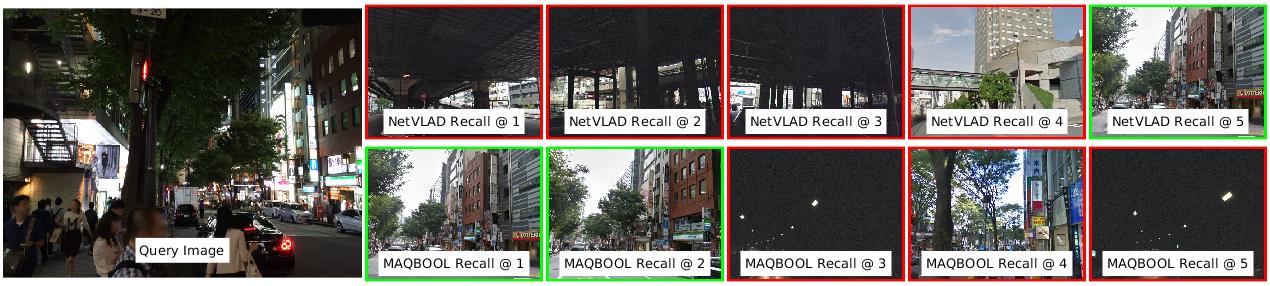}
	}
	\captionsetup[subfigure]{width=.93\linewidth}
	\subfloat[Recall test on Tokyo 24/7 dataset at a features dimension of 4096.] {\label{fig:results-thumb-tokyo2tokyo-4096} 
		\includegraphics[width=.99\linewidth]{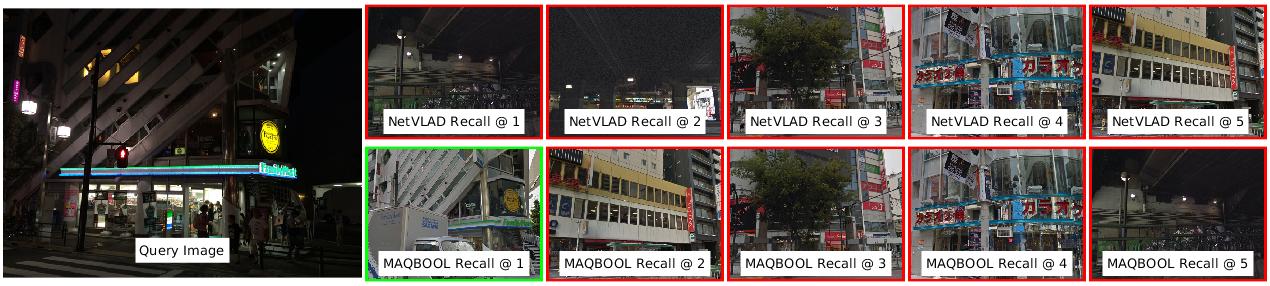}
	}
	\captionsetup[subfigure]{width=.93\linewidth}
	\subfloat[Recall test on Pitts250k dataset at a features dimension of 512.] {\label{fig:results-thumb-pitts2pitts-512} 
		\includegraphics[width=.99\linewidth]{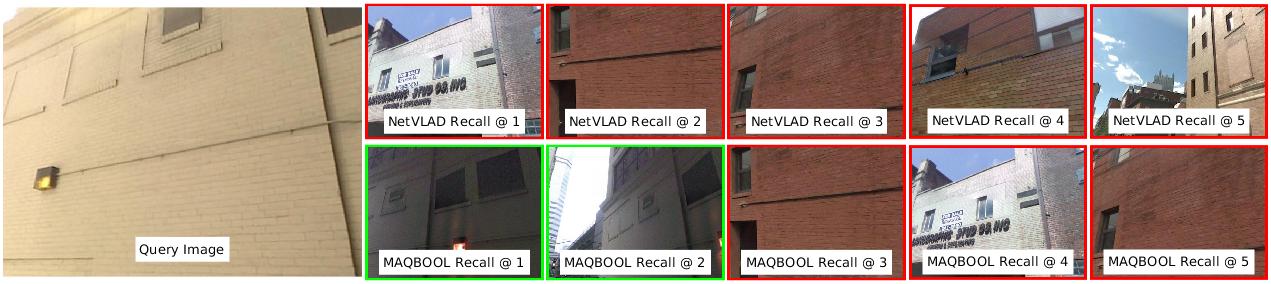}
	}
	\captionsetup[subfigure]{width=.93\linewidth}
	\subfloat[Recall test on Pitts250k dataset at a features dimension of 4096.] {\label{fig:results-thumb-pitts2pitts-4096}
		\includegraphics[width=.99\linewidth]{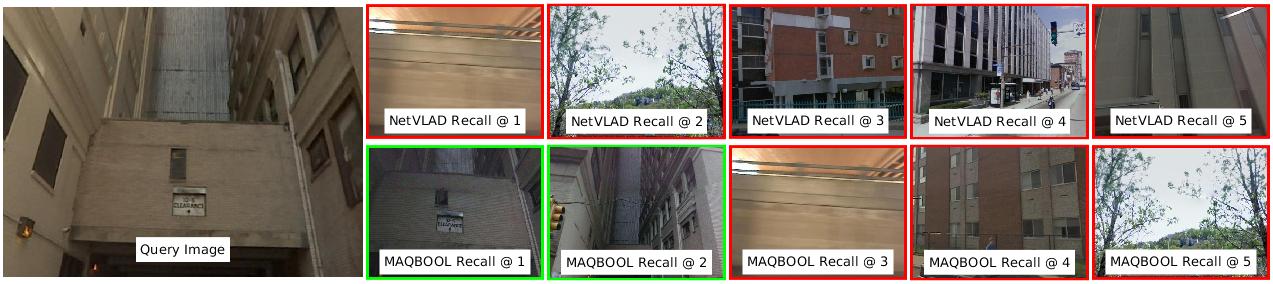}
	}
	\caption{MAQBOOL vs. NetVLAD: (a) and (b) show the image retrieval compared to NetVLAD on the Tokyo 24/7 dataset with feature dimensions 512 and 4096. (c) and (d) show the recall on the Pittsburgh dataset with feature dimensions 512 and 4096.}
	\label{fig:results-thumb} 
\end{figure*}

The top five recall results tested on the Pittsburgh and Tokyo 24/7 datasets are shown in Fig. \ref{fig:results-thumb}. It is shown in Fig. \ref{fig:results-thumb-tokyo2tokyo-512} and \ref{fig:results-thumb-tokyo2tokyo-4096} that NetVLAD fails to retrieve the nearest match with the query in the first five places of the Tokyo 24/7 dataset, while MAQBOOL successfully adjusts the distances of the retrieved images and re-ranks the closest match to the first position. Similarly, Fig. \ref{fig:results-thumb-pitts2pitts-512} and \ref{fig:results-thumb-pitts2pitts-4096} show the robustness of our proposed system compared to NetVLAD when tested on the Pittsburgh dataset at feature dimensions of 512 and 4096, respectively.

\subsection{MAQBOOL Representation}
In visual place recognition, the standard baseline is to increase the feature dimension. Our work proved that we could make it better by adding probabilistic information. SLAM systems do not recognize the place at 30/60 frames per second (FPS), instead they detect loop closure at nearly 1 second intervals. We perform landmarks-based verification that takes longer than one second. There is always a trade-off between speed and accuracy. Edge Boxes takes nearly 0.37 second in MATLAB. We used $n=50$ regions in this work, which takes an additional 0.87 sec. Depending on the application, we can reduce the $n$ regions. For instance, the system takes only 0.13 second instead of 0.87 sec for processing five boxes. 

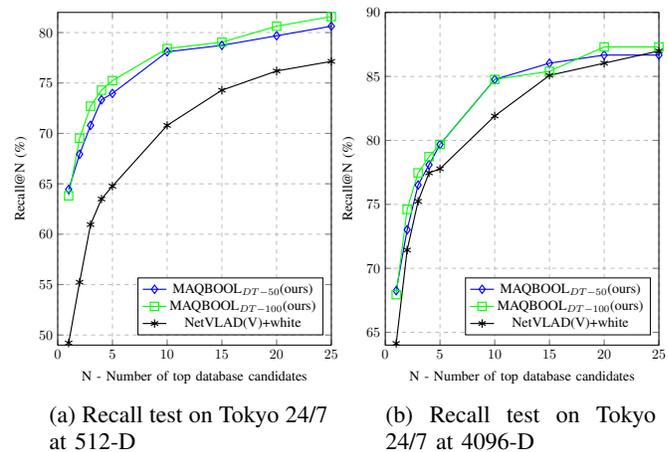
\begin{figure}[!t] 
	\centering
	\raggedright	
	\captionsetup[subfigure]{width=.85\linewidth}
	\subfloat[Recall test on Tokyo 24/7 at 512-D] 
	{	\label{fig:pitts2tokyo-512} 
		\resizebox{.45\linewidth}{!}{
			\pgfplotsset{ymax=82, ymin= 49}
			\begin{tikzpicture}[trim axis left]%
				\begin{axis}
					[curve plot style]
					\addplot [color=blue, mark=diamond, mark size=3pt, line width=0.8pt]
					table {data/vd16_pitts30k_to_tokyo247_maqbool_DT_50_512.dat}; %
					\addplot [color=green, mark=square, mark size=3pt, line width=0.8pt]
					table {data/vd16_pitts30k_to_tokyo247_maqbool_DT_100_512.dat}; %
					\addplot [color=black, mark=asterisk, mark size=3pt, line width=0.8pt]
					table {data/vd16_pitts30k_to_tokyo247_netvlad_512.dat};
					\legend{MAQBOOL$_{DT-50}$(ours), MAQBOOL$_{DT-100}$(ours),NetVLAD(V)+white}
				\end{axis}
			\end{tikzpicture}
		}
	}
	\captionsetup[subfigure]{width=.85\linewidth}
	\subfloat[Recall test on Tokyo 24/7 at 4096-D] 
	{\label{fig:pitts2tokyo-4096}
		\resizebox{.45\linewidth}{!}{
			\pgfplotsset{ymax=90, ymin= 64}
			\begin{tikzpicture}[trim axis left]%
				\begin{axis}
					[curve plot style]
					
					\addplot [color=blue, mark=diamond, mark size=3pt, line width=0.8pt]
					table {data/vd16_pitts30k_to_tokyo247_maqbool_DT_50_4096.dat}; %
					\addplot [color=green, mark=square, mark size=3pt, line width=0.8pt]
					table {data/vd16_pitts30k_to_tokyo247_maqbool_DT_100_4096.dat}; %
					\addplot [color=black, mark=asterisk, mark size=3pt, line width=0.8pt]
					table {data/vd16_pitts30k_to_tokyo247_netvlad_4096.dat};
					\legend{MAQBOOL$_{DT-50}$(ours), MAQBOOL$_{DT-100}$(ours),NetVLAD(V)+white}
				\end{axis}
			\end{tikzpicture}
		}
	}
	\caption{MAQBOOL performance comparison with NetVLAD on Tokyo 24/7 dataset. Their retrieval performances are evaluated using the same VGG model, which is trained on the Pitts30k dataset.}
	\label{fig:pitts2tokyo}
\end{figure}

Let's assume we are utilizing the SLAM system on Mars, and we choose the model trained in some datasets. In that case, vanilla NetVLAD cannot perform with high accuracy. Fig.  \ref{fig:pitts2tokyo} shows the MAQBOOL performance on the Tokyo 24/7 datasets with 512-D and 4096-D features compared with the state-of-the-art NetVLAD. Both use the same model trained on the Pittsburgh 30k dataset and tested on the Tokyo 24/7 dataset. We achieved notable improvement at low-dimensional (512-D) feature-based recall, comparable with the 4096-D NetVLAD. 

\begin{table*}[!t]
	\caption{MAQBOOL performance comparison with APANet and NetVLAD at 512-D.}
	\centering
	\setlength{\tabcolsep}{1.3em} %
	\renewcommand{\arraystretch}{1.2}%
	\begin{tabular}{|c|c|c|c|c|c|c|c|}
		\hline
		\multirow{2}{*}{Method}                           & \multirow{2}{*}{Whitening} & \multicolumn{3}{c|}{Tokyo 24/7} & \multicolumn{3}{c|}{Pitts250k-test}                                                                                                               \\ \cline{3-8}
		
		&                            & \multicolumn{1}{c|}{Recall@1}   & \multicolumn{1}{c|}{Recall@5}       & \multicolumn{1}{c|}{Recall@10} & \multicolumn{1}{c|}{Recall@1} & \multicolumn{1}{c|}{Recall@5 } & Recall@10 \\
		\hline
		\hline

		\multirow{2}{*}{Sum pooling}                      & PCA whitening              & 44.76                           & 60.95                               & 70.16                           & 74.13                         & 86.44                          & 90.18     \\
		
		& PCA-pw                     & 52.70                            & 67.30                                & 73.02                          & 75.63                         & 88.01                          & 91.75    \\
		\hline
		\multirow{2}{*}{NetVLAD \cite{arandjelovic2016netvlad} } & PCA whitening              & 56.83                              & 73.02                               & 78.41                          & 80.66                         & 90.88                          & 93.06     \\
		
		& PCA-pw                     & 58.73                           & 74.6                                & 80.32                           & 81.95                         & 91.65                          & 93.76    \\
		\hline
		\multirow{2}{*}{APANet \cite{zhu2018attention} }  & PCA whitening              & 61.90                            & 77.78                               & 80.95                          & 82.32                         & 90.92                          & 93.79     \\
		
		& PCA-pw                     & 66.98                           & \textbf{80.95}                               & 83.81                          & 83.65                         & 92.56                          & 94.70      \\
		
		\hline
		\multirow{2}{*}{MAQBOOL (Ours)}                   
		& PCA whitening + DT-50      & 68.25                           & 79.37                               & 83.49                           & 85.45           & 92.62                          
		& 94.58     \\
		& PCA whitening + DT-100     & \textbf{69.21}                        & 80.32           & \textbf{84.44}                        & \textbf{85.46}                     & \textbf{92.77}                          & \textbf{94.72}    \\

		\hline
	\end{tabular}
	\label{table-1}
\end{table*}

\subsection{Comparison with Power Whitening PCA}
While maintaining the same baseline of PCA whitening followed by NetVLAD, MAQBOOL outperforms NetVLAD as well as APANet, as shown in Table \ref{table-1}. APANet introduces an additional PCA power whitening concept on different block types and produces better performance than NetVLAD, but by keeping the default PCA whitening, our simplest model delivers better results than APANet. We observe that by increasing the tree size, there is a significant improvement in the accuracy at a high dimension of 4096. Moreover, for a decision tree dimension of 50, MAQBOOL achieves good results compared to APANet on the Tokyo 24/7 and Pittsburgh datasets, as shown in Table \ref{table-1}. 

\subsection{Ablation Study}
As mentioned in the previous section, we use a decision tree model at the PDL. We choose the decision tree and Gaussian probability models for use in the PDT layer, and we discuss them as follows.

\subsubsection{Decision tree model}
We trained the default model based on '100' tree size. For the ablation study, we trained a decision tree model of size '50'. We found that by reducing the tree size, the system performance is almost similar. MAQBOOL with a decision tree size of '100' works slightly better at a higher dimension (4096-D) than MAQBOOL with a decision tree size of '50', as shown in Fig. \ref{fig:4plots}. Furthermore, We find that the tree size of '50' has a better recall than NetVLAD and power-whiteing-based APANet. 
We choose ground truth data of the Tokyo Time Machine validation set to create the model. However, we observed that the model works better than NetVLAD if created using small datasets, such as the Oxford 5k \cite{oxford5k} and Paris \cite{paris6k} datasets. These datasets have 55 query images. However, prediction models based on these datasets show similar performance.

\subsubsection{Gaussian probability model}
The Gaussian probability model is also a popular choice in regression studies. We observe that it has a similar performance with the decision tree of size '50'.

\section{Conclusion}
\label{sec.Conclusion}
In this paper, we introduce MAQBOOL to improve the accuracy of image retrieval results without retraining a new deep learning model, for better visual place recognition. We elevate essential regions at spatial layers and probabilistically verify the image correspondence efficiently. Our MAQBOOL approach intelligently processes the high-level layer to produce more-reliable top matches than the current state-of-the-art. Without any further training or introducing additional sensors or pieces of ground truth information to the system, our framework outperforms PCA power whitening on APANet, and on NetVLAD. {Our method achieves good accuracy on low-dimensional features (i.e., 512-D) with more reliable candidates, which makes it useful in general SLAM applications for loop closure detection.}
\section{Appendix}
The primary motivation behind this work is to make the multiagent SLAM systems efficacious towards new deployment. Image retrieval is a key part of not only the SLAM system but also of data analytics. This paper suggests an intelligent way to use the top landmarks for better place recognition. If we observe the recall rate of NetVLAD while tested on challenging datasets such as the Tokyo 24/7 dataset\cite{Torii2018}, Tokyo Time Machine, and Pittsburgh dataset\cite{Torii2015}. We found that model should be trained using the same dataset to achieve a good recall rate. Moreover, the top 10-25\% candidates from the database have recall rates with accuracy 90\% and above. Generally, SLAM systems take the top first candidate from the retrieval for loop closure detection. It means we cannot emphasize the direct usage of NetVLAD for single-loop-closure-based multiagent system \cite{bhutta2020loopbox}.

\subsection{Evaluation on Oxford Building and Paris Building datasets}
\label{app-extended-results}

\begin{figure}[!t] 
	\raggedright	
	\captionsetup[subfigure]{width=.93\linewidth}
	\subfloat[ Recall test at a features dimension of 512.] {\label{fig:results-thumb-tokyo2tokyo-512} 
		\includegraphics[width=.99\linewidth]{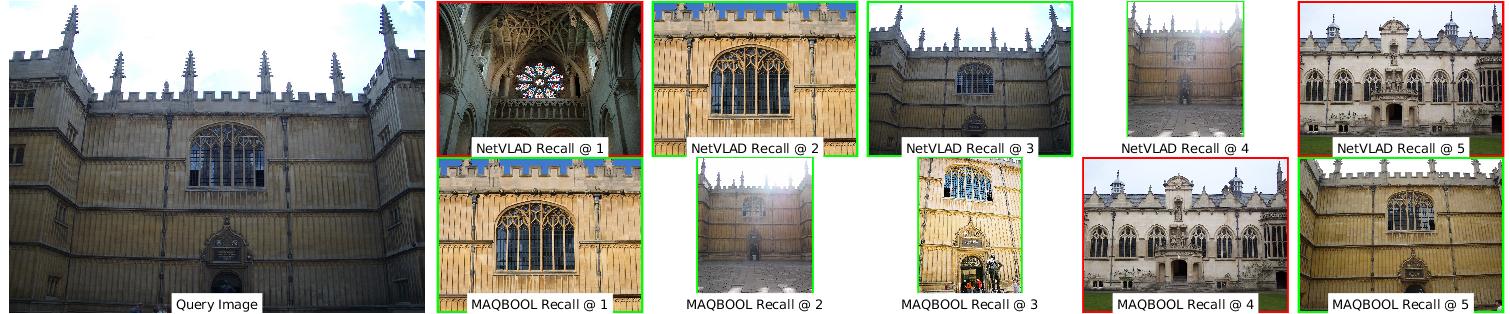}
	}
	\captionsetup[subfigure]{width=.93\linewidth}
	\subfloat[ Recall test at a features dimension of 512.] {\label{fig:results-thumb-tokyo2tokyo-4096} 
		\includegraphics[width=.99\linewidth]{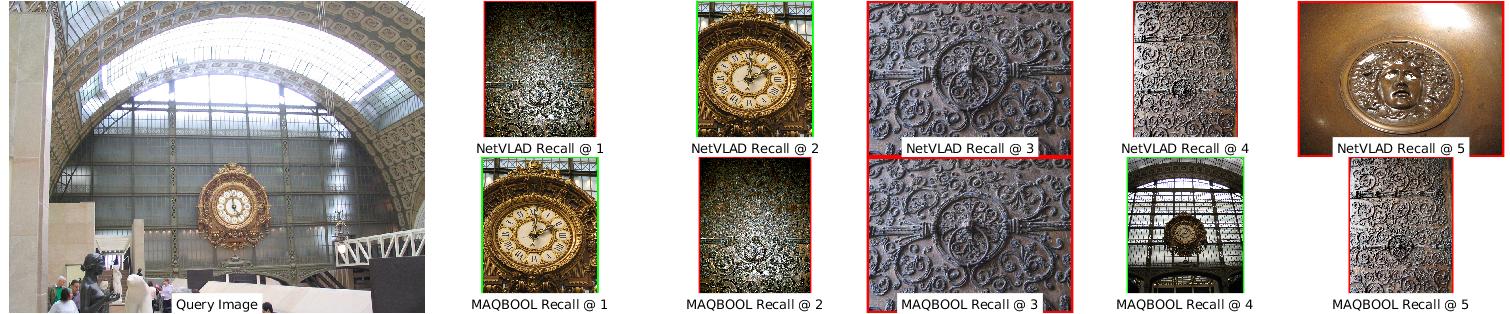}
	}
	
	\caption{ MAQBOOL vs. NetVLAD: (a) Evaluation is carried out of the Oxford 5k \cite{oxford5k} dataset  (b) Evaluation is carried out of the Paris 6k \cite{paris6k} dataset.}
	\label{fig:oxfpar} 
\end{figure}

In Fig. \ref{fig:oxfpar}, NetVLAD failed to put the right matches at the first position for each query image from Oxford 5k building and Paris 6k building datasets. Our work at the lower feature dimension, i.e., 512-D, successfully places the right matches to the first position. That makes it more robust towards using any mapping system. For the localization system, the first recall is very important for a complete global mapping optimization.

\subsection{Recall Improvement at a Lower Features Dimension}
\begin{figure}[!h]
	\begin{subfigure}[b]{.40\linewidth}
		\centering
		{\includegraphics[width=.99\linewidth]{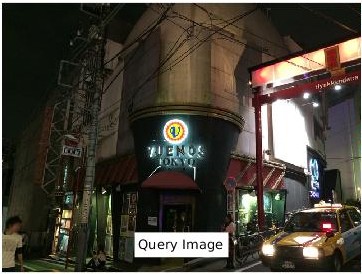}}
		\caption{Query Image.}\label{fig:1a}
	\end{subfigure}%
	\begin{subfigure}[b]{.59\linewidth}
		\centering
		\subcaptionbox{\label{fig:1d}NetVLAD recall at 512-D.}
		{\includegraphics[width=.99\linewidth]{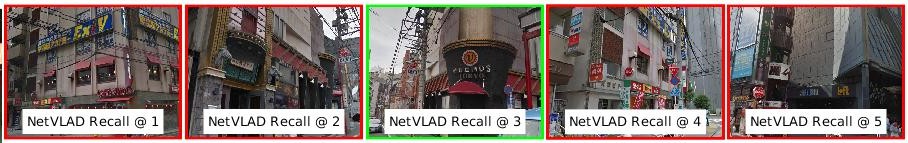}}
		
		\subcaptionbox{\label{fig:1e}NetVLAD recall at 4096-D.}
		{\includegraphics[width=.99\linewidth]{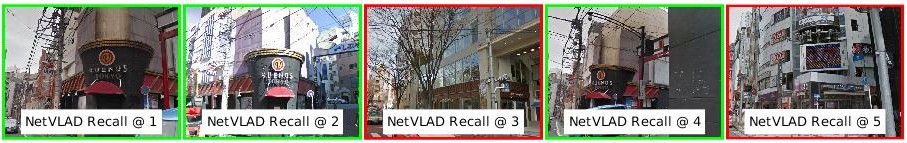}}%
		
		\subcaptionbox{\label{fig:1g}MAQBOOL recall at 512-D.}
		{\includegraphics[width=.99\linewidth]{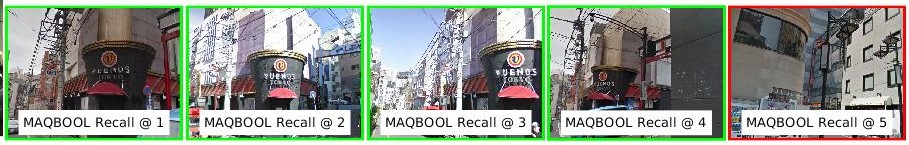}}
	\end{subfigure}%
	\caption{ Extended results of Fig. 1(a). Query image is taken from Tokyo 24/7 dataset. (b), (c) and (d) show the first five recalls from the database.}
	\label{app-1} 
\end{figure}

\begin{figure}[!h]
	\begin{subfigure}[b]{.40\linewidth}
		\includegraphics[width=.99\linewidth]{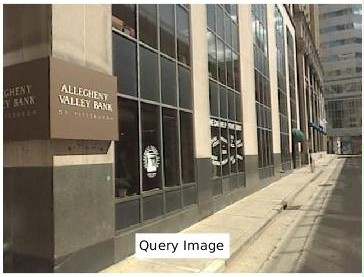}
		\caption{Query Image}\label{fig:1a}
	\end{subfigure}%
	\begin{subfigure}[b]{.59\linewidth}
		\centering
		\subcaptionbox{\label{fig:1d}NetVLAD recall at 512-D.}
		{\includegraphics[width=.99\linewidth]{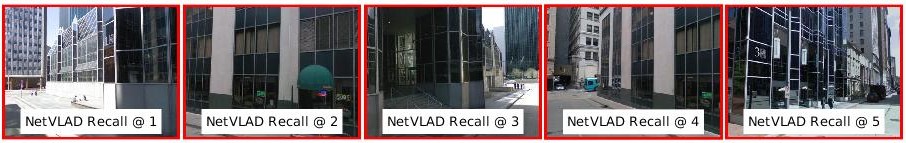}}
		
		\subcaptionbox{\label{fig:1e}NetVLAD recall at 4096-D.}
		{\includegraphics[width=.99\linewidth]{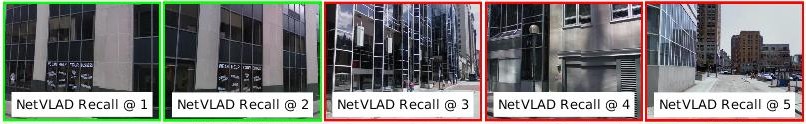}}%
		
		\subcaptionbox{\label{fig:1g}MAQBOOL recall at 512-D.}
		{\includegraphics[width=.99\linewidth]{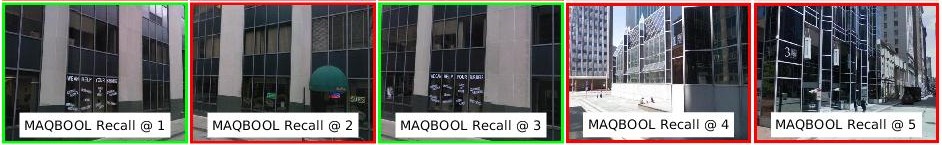}}
	\end{subfigure}%
	\caption{Extended results of Fig. 1(b). Query image is taken from Pittsburgh dataset. (b), (c) and (d) show the first five recalls from the database.}
	\label{app-2}
\end{figure}

Fig. \ref{fig:a} show the quantitative results of MAQBOOL at 512-D compared to NetVLAD at 512-D and 4096-D. In Fig. \ref{app-1} and \ref{app-2}, we show qualitative results of our approach compared with the recall of NetVLAD. We found that MAQBOOL at 512-D outperformed the NetVLAD at 4096-D for the low light images. 
For the query image shown in Fig. \ref{app-1}, our MAQBOOL at 512-D with DT-50, not only correct the first match as NetVLAD at 4096-D did, but also it brings the correct match at the third position.

\ifCLASSOPTIONcaptionsoff
  \newpage
\fi

\bibliographystyle{IEEEtran}
\bstctlcite{BSTcontrol}
\bibliography{root}

\end{document}

%% file: tikz_styles.tex
\usepackage{pgfplots}

\pgfplotsset{
	curve plot style/.style={
		xlabel={N - Number of top database candidates},
		xmin=0, xmax=25,
		xtick={0,5,10,15,20,25},
		ytick={45,50,55,60,65,70,75,80,85,90,95,100},
		legend pos=south east,
		ymajorgrids=true,     xmajorgrids=true,
		grid style=dashed,
		ylabel={Recall@N (\%)},
		ylabel style={yshift=-.3cm},
		width=8.5cm,
		height=10cm,
	},
}